%% file: Formatting-Instructions-LaTeX-2022.tex
\def\year{2022}\relax
\documentclass[letterpaper]{article} 
\usepackage{aaai22}  
\usepackage{times}  
\usepackage{helvet}  
\usepackage{courier}  
\usepackage[hyphens]{url}  
\usepackage{graphicx} 
\usepackage{natbib}  
\usepackage{caption} 
\DeclareCaptionStyle{ruled}{labelfont=normalfont,labelsep=colon,strut=off} 
\usepackage{algorithm}
\usepackage{algorithmic}
\usepackage{amsfonts}
\usepackage{amssymb}
\usepackage{mathtools}
\usepackage{mathbbol}
\usepackage{arydshln}
\usepackage{newfloat}
\usepackage{listings}
\floatstyle{ruled}
\newfloat{listing}{tb}{lst}{}
\floatname{listing}{Listing}
\usepackage[switch]{lineno}
\usepackage{subfigure}
\usepackage{multirow}
\usepackage{booktabs}
\title{Towards Off-Policy Reinforcement Learning for Ranking Policies \\ with Human Feedback}

\author{
    Teng Xiao, Suhang Wang}
\affiliations{
     The Pennsylvania State University\\
    \{tengxiao, szw494\}@psu.edu
}

\begin{document}

\maketitle

\begin{abstract}
Probabilistic learning to rank (LTR) has been the dominating approach for optimizing the ranking metric, but cannot maximize long-term rewards. Reinforcement learning models have been proposed to maximize user long-term rewards by formulating the recommendation as a sequential decision-making problem, but could only achieve inferior accuracy compared to LTR counterparts, primarily due to the lack of online interactions and the characteristics of ranking. In this paper, we propose a new off-policy value ranking (VR) algorithm that can simultaneously maximize user long-term rewards and optimize the ranking metric offline for improved sample efficiency in a unified Expectation-Maximization (EM) framework.  We theoretically and empirically show that the EM process guides the leaned policy to enjoy the benefit of integration of the future reward and ranking metric, and learn without any online interactions. Extensive offline and online experiments demonstrate the effectiveness of our methods. 
\end{abstract}

\input{Introduction}

\input{Notations_Formulation}
\input{Method}
\input{Experiment}

\input{Conclusion}

\section*{Acknowledgments}
This research is supported by, or in part by, the National Science Foundation (NSF) under grant IIS-1909702 and IIS-1955851, and Army Research Office (ARO) under grant W911NF21-1-0198.

\bibliography{Reference}

\end{document}

%% file: Introduction.tex
\section{Introduction}
\label{sec:intro}
With the advances of deep learning, various deep learning based recommender systems (RS) have been proposed to capture the dynamics of  user preferences. However, they are mainly based on the  probability ranking principle~\cite{robertson1977probability} that the optimal ranking should rank items in terms of probability of relevance to a user. Typically, these methods are trained based on  \textit{Maximum Likelihood Estimation} (MLE) on the past logged feedbacks with supervised pointwise~\cite{hu2008collaborative}, pairwise~\cite{rendle2012bpr,burges2005learning}, or listwise~\cite{cao2007learning}  loss functions. Probability learning to rank  (LTR) is  one of the guiding technical principles behind the optimization of ranking models in  information retrieval.  Recently, reinforcement learning (RL) is  gaining a lot of attraction in  recommender system~\cite{DBLP:conf/wsdm/ChenBCJBC19,DBLP:conf/kdd/ZhaoZDXTY18,DBLP:conf/wsdm/ZouXDZB0NY20,xiao2021general,xin2020self}. Different from  probability LTR which focuses on picking a  model under the immediate feedback distribution, RL in general focuses on learning policy that takes actions in a dynamic environment so as to maximize the long-term reward. 

Despite its attraction, optimizing the RL objective under the real recommendation scenario has the following  challenges. 
\textbf{(1)} Missing online interactions. In contrast to probability ranking methods, training an optimal RL algorithm requires a large number of online interactions with the environments (real users)~\cite{fujimoto2019off}, which is impractical as it could hurt user experiences when the RL agent is not well-trained, especially at the beginning of training.  Without online interactions,
RL algorithms such as the Q-learning~\cite{haarnoja2017reinforcement} and Actor-Critic~\cite{haarnoja2018soft,lillicrap2015continuous} suffer from the overestimation issue~\cite{fujimoto2019off,kumar2020conservative}, a phenomenon that unseen state-action pairs are erroneously estimated to have unrealistic values if there is no online interaction with real users, which is also verified by our preliminary analysis in \S~\ref{sec:RLTR}. \\
\textbf{(2)} Lack of characteristics of ranking. Off-policy RL methods, such as one-step Q-learning~\cite{watkins1992q} and
variants of deep Q networks (DQN)~\cite{schaul2016prioritized}, enjoy the advantage of learning from any samples from the same environment (i.e., off-policy learning). However,  as shown in our analysis \S~\ref{sec:RLTR}, these methods can be seen as a regression problem that focuses on directly estimating optimal state-action value. However, in recommendation, we are more interested in the relative ranking of the actions (items) given user states and optimizing the ranking metric instead of the absolute values of actions, which is the main difference between recommendation and the robotics domain where only one optimal action is needed at each time step. \\
\textbf{(3)} Partial and sparse feedback.  In RS, click data, which is called implicit data, are easy to collect because they represent the behavior logs of the users. As the implicit data is not the explicit feedback of the users’ preferences, one cannot know whether unclicked  feedback is negative feedback or unlabeled positive feedback. Directly considering the reward of  unclicked  feedback as negative would  lead to a suboptimal ranking algorithm. As shown in~\cite{DBLP:conf/wsdm/ChenBCJBC19}, standard off-policy learning methods do not take into account unobserved feedbacks. 
Motivated by the discussions above, in this paper, we investigate the problem of learning a ranking algorithm that can simultaneously maximize the future reward and optimize the ranking metric  based on logged feedbacks.  It is a challenging problem as: (i) There is a huge optimization gap between MLE and RL since they are totally different learning paradigms for ranking as we discussed above;  (ii) In practice, there is no access to the reward for every state-action pair. In other words, we only observe \textit{partial} and \textit{sparse} reward  on the logged feedbacks; and (iii) RL  is prone to be overestimated~\cite{levine2020offline}, i.e., they will overestimate the values of the actions, resulting in bad performance. The overestimation is much more severe without online interactions~\cite{kumar2020conservative}. 

To address these challenges, we propose a novel learning framework to conduct ranking  with rewards based on the Expectation-Maximization (EM) principle. Our framework can unify the MLE and RL, where both the RL teacher and the MLE student help each other in a closed-loop and gives extra power for knowledge distillation between reward  and  feedbacks. To solve the problem of partial and sparse rewards,  we  propose reward extrapolation and ranking regularization to improve our EM framework. 
We extend the EM framework to sequential settings and propose a novel algorithm named  off-policy value ranking (VR), for maximizing long-term rewards. VR optimizes the ranking metric  by learning the  relative ranking of  value function instead of estimating the absolute values. The main contributions are:\\
(1) We propose an EM framework  that can learn from the sparse and partial reward signal through unifying the probability and reinforcement ranking principles.\\
(2) We extend our framework to sequential settings and propose an off-policy ranking algorithm that maximizes long-term rewards without online interactions and achieves better ranking performance both empirically and theoretically.\\
(3) Experiments on two datasets with three state-of-the-art backbones and the simulated online experiments on the RecSim environment show the effectiveness of our framework.

%% file: Notations_Formulation.tex
\section{Preliminaries}
\subsection{Notations and Problem Definition}
Assume we have a set of users $u \in \mathcal{U}$, a set of items $i \in \mathcal{I}$. Following previous work~\cite{DBLP:conf/wsdm/ChenBCJBC19,DBLP:conf/sigir/ZhouDC0RTH020}, we can translate the task of recommendation into a markov decision process (MDP): $\left(\mathcal{S}, \mathcal{A}, {P}, R, \rho, \gamma\right)$ where \\
$\bullet \mathcal{S}$: a  space describing  user states: ${s}_{t}=\left(i_{1}, i_{2}, \cdots, i_{t}\right)$, where $i_t$ denotes the item interacted (clicked) by users; \\
$\bullet \mathcal{A}:$ a discrete action space which are recommending items;\\
$\bullet {P}: \mathcal{S} \times \mathcal{A} \times \mathcal{S} \rightarrow \mathbb{R}$  is the   state transition probability.\\
$\bullet R$: $\mathcal{S} \times\mathcal{A} \rightarrow \mathbb{R}$ denotes the reward, where $r({s}_{t}, {a}_{t})=r_{t}$ is the received  reward  by performing action ${a}_{t}$ at user state ${s}_{t}$;\\ 
$\bullet \rho({s}_1)$ is the initial distribution and $\gamma$ is the discount factor; \\
Formally, we seek a policy $p_{\theta}(\cdot | {s}_{t})$ which translates the user state ${s} \in \mathcal{S}$ into a distribution of the action space,  so as to maximize  expected cumulative rewards as follows:  
\begin{linenomath}
\small
\begin{align}
\max _{{\theta}} \mathcal{J}({\theta})=\mathbb{E}_{\tau \sim p_{\theta}(\tau)}[R(\tau)], R(\tau)={\sum}_{t=1}^{T} \gamma^{t} r\left({s}_{t}, a_{t}\right), \label{Eq:RL1}
\end{align}
\end{linenomath}
where the expectation is taken over the trajectory $\tau=\left({s}_{1}, a_{1}, \cdots {s}_{T}\right)$ obtained by acting policy step by step: ${s}_{1} \sim \rho({s}_1), a_{t} \sim p_{\theta}\left(a_t | {s}_{t}\right)$ and ${s}_{t+1} \sim p\left({s}_{t+1}|{s}_{t}, a_{t}\right)$. 
Generally, the agent cannot interact with the user  as we are only provided with a  logged  dataset,  where the action $a_{t}$ is chosen by a historical logging (behavior) policy $p_{\text{b}}=p_{\text{b}}(a_t | {s}_t)$.  The past \textit{implicit feedbacks} (only positive feedbacks)  can  be represented as $\mathcal{D}=\{\{{s}_{t}^{i}, a_{t}^{i}, r_{t}^{i},{s}_{t+1}^{i}\}_t\}_{i=1}^{N}$, where $N$ is the total number of data. For brevity, we omit  superscript $i$  in what follows. 
With the definition above, our studied problem is formally defined as: \textit{Given  past interactions $\mathcal{D}$, our goal is to learn the recommendation policy $p_{\theta}(a|s)$ which can maximize  cumulative future reward, i.e.,~Eq.~(\ref{Eq:RL1}) for RS.}
 \begin{figure*}[t]
\centering
  \subfigure{
    \includegraphics[width=0.221\textwidth]{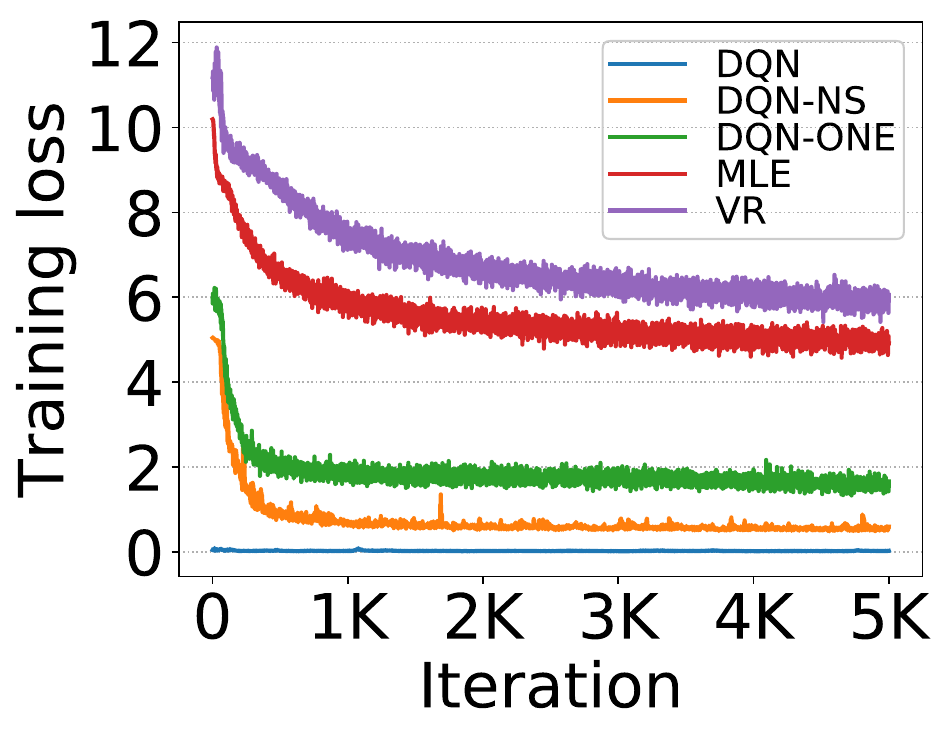}}
  \subfigure{
    \includegraphics[width=0.23\textwidth]{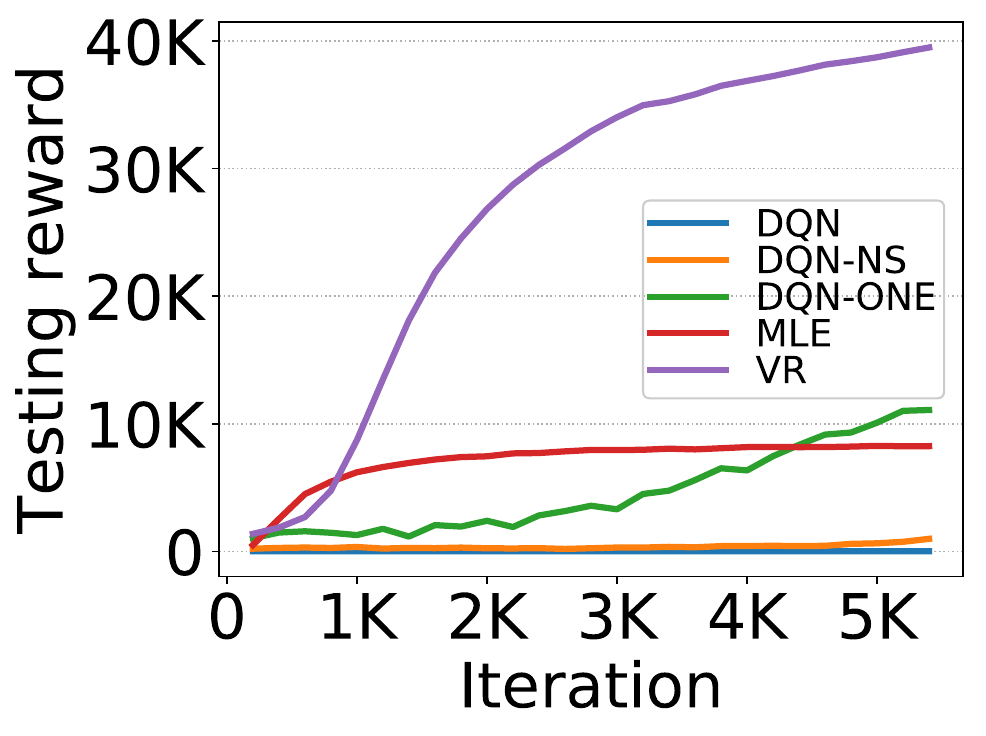}}
          \subfigure{
    \includegraphics[width=0.233\textwidth]{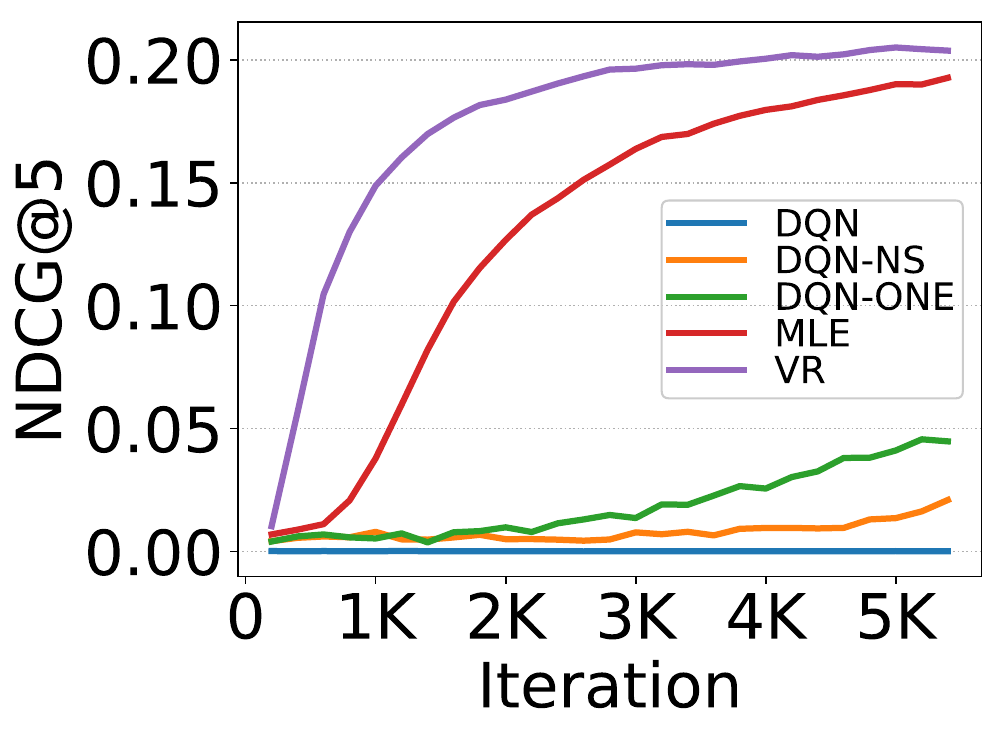}}
        \subfigure{
    \includegraphics[width=0.233\textwidth]{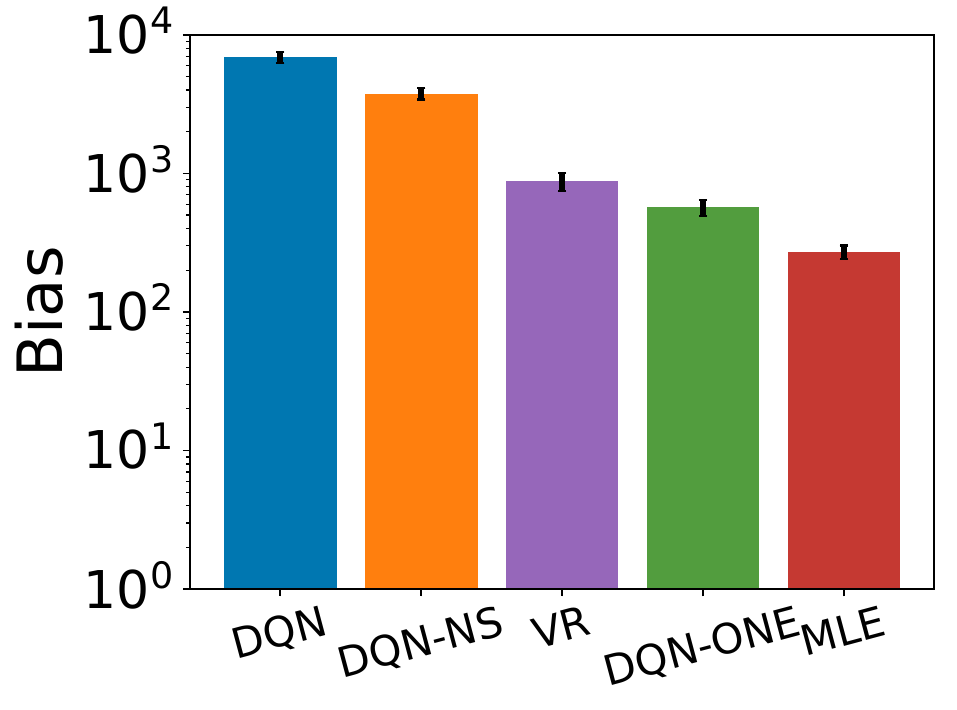}}
        \vskip -1.5em
    \caption{Training curves and overestimation bias of MLE, our VR, DQN, and two variants of DQN.}\label{fig:pre}
    \vskip -1em
\end{figure*}

\subsection{Explorations on MLE- \& RL-based RS}
\label{sec:RLTR}
\textbf{MLE}: Typically,  current   sequential recommendation models~\cite{tang2018personalized,kang2018self,xiao2019hierarchical}  rely on the probability ranking principle and are trained to approximate the ranking metric based on MLE. Specifically,  MLE  maximizes the log-likelihood of  logged feedbacks   as follows:
\begin{linenomath}
\small
\begin{align}
\mathcal{L}_{\text{MLE}}({\psi})=\mathbb{E}_{p_{\text{b}}\left(a_{t}| s_{t}\right) }\left[\log p_{{\psi}}(a_t| s_t)\right] \label{MLE}.
\end{align}
\end{linenomath}
Note that the actions are clicked or purchased items since we only observe positive feedbacks. Although we do not know $p_{\text{b}}\left(a_{t}| s_{t}\right)$, we can get its samples from logged feedbacks. For listwise models~\cite{cao2007learning}, $p_{\psi}(a_t| s_t)$ is  parameterized neural networks~\cite{kang2018self} with the last softmax layer. Besides the listwise loss,  pairwise loss such as Bayesian personalized ranking~\cite{rendle2012bpr} also have shown to be effective in ranking. Although  MLE can naturally approximate the ranking metric~\cite{xia2008listwise}, it cannot maximize the  long-term reward of users. \\
\noindent{}\textbf{RL}: Various RL-based recommendation algorithms have been proposed~\cite{DBLP:conf/www/ZhengZZXY0L18,DBLP:conf/wsdm/ZouXDZB0NY20,zhao2018ac}. Generally, these algorithms try to minimize the following  temporal difference (TD) error as:
\begin{linenomath}
\small
\begin{align}
\mathcal{L}_{TD}({\phi})=\left(Q_{{\phi}}\left({s}_{t}, a_{t}\right)-r_t-\gamma Q_{\bar{\phi}}\left({s}_{t+1}, \pi\left({s}_{t+1}\right)\right)\right)^{2}, \label{TD}
\end{align}
\end{linenomath}
where $Q_{{\phi}}\left({s}_{t}, a_{t}\right)$ is a learning state-action value function and $\pi\left({s}_{t+1}\right)=\arg \max _{a} Q_{\bar{\phi}}\left({s}_{t+1}, a\right)$. The $\bar{\phi}$ corresponds to the frozen weights in the target network, and is updated at fixed intervals~\cite{lillicrap2015continuous}. While Q-learning is an effective off-policy algorithm in robotics, we argue that it is not suitable for the ranking problem in RS as: \textit{Q-learning can essentially be viewed as a supervised regression model for predicting the true value; while the relative ranking of value is more important in recommendation}. Suppose the discount factor $\gamma=0$, then the TD error in Eq.~(\ref{TD}) becomes:
\begin{linenomath}
\small
\begin{align}
\hat{\mathcal{L}}_{TD}({\phi})=\mathbb{E}_{\left({s}_{t}, a_{t}, r_t, {s}_{t+1}\right) \sim \mathcal{D}}\left(Q_{{\phi}}\left({s}_{t}, a_{t}\right)-r_t\right)^{2},  \label{ONETD}
\end{align}
\end{linenomath}
which is exactly the mean squared error (MSE) that estimates reward value. However, as shown by previous  works~\cite{liang2018variational},  MSE  is not a good proxy for the top-N ranking metric. Besides, when  $\gamma \neq 0$, since the TD loss in Eq.~(\ref{TD}) selects next actions via $\max _{a} Q_{\bar{\phi}}\left({s}_{t+1}, a\right)$, it  suffers from the overestimation~\cite{levine2020offline}: $\max _{a} Q_{\bar{\phi}}\left({s}_{t+1}, a\right)$ of  out-of-distribution actions is too big to be correct.\\
\noindent{}\textbf{Empirical analysis}. To verify our discussion above, we 
conduct preliminary analysis to understand the ranking, partial feedbacks and overestimation issues of Q-learning in RS. We compare the Q-learning (DQN) and supervised MLE (list-wise) with our proposed VR on YooChoose (see \S~\ref{Overall performance} for settings).  We also consider two variants of DQN, i.e., DQN-NS and DQN-ONE. DQN-NS uniformly samples unseen items to provide negative (-1) rewards. DQN-ONE, i.e., Eq.~(\ref{ONETD}) is the one-step DQN which sets the discount factor $\gamma$ in DQN to zero. We also compute the overestimation bias over the test set $\mathcal{D}_{t}$ as: $\frac{1}{|\mathcal{D}_{t}|} \sum_{(s,a)\in\mathcal{D}_{t}}(\max (Q_{\phi}(s, a)-V(s), 0))^{2}$, where $V(s)$  denotes the discounted sum of rewards. From Fig.~\ref{fig:pre}, we have some important findings.\\
 (1) We can find the ranking performance of  DQN and DQN-NS is extremely bad, which empirically verifies our discussion  in \S~\ref{sec:RLTR} that the TD-based DQN is not suitable for learning ranking policy offline due to  the large \textit{overestimation bias} and lack of \textit{ranking characteristics}. (2) DQN-ONE outperforms DQN and DQN-NS, which shows that DQN and DQN-NS can not benefit from future rewards at all due to the overestimation bias. (3) DQN-NS reduces this bias to some extent but still suffer it severely, which demonstrates that incorporating negative evidence is important when training the RL algorithm for the \textit{partial feedbacks} in RS, and this is exactly one of our motivations for adding ranking regularization in our VR. Besides the Q-learning based RL, \cite{DBLP:conf/wsdm/ChenBCJBC19} applies off-policy policy gradient to learn from logged feedbacks for RS. Specifically, they utilize the  importance sampling technique~\cite{munos2016safe} to conduct off-policy estimation. While this method can learn from logged feedback as an off-policy way, it is still  biased and  has too large variance to be effective~\cite{levine2020offline}.

%% file: Method.tex
\section{Learning to Rank with  Rewards}
\label{sec:L2R}
Based on the analysis above, we summarize the limitations of current RS algorithms: (1) The probability ranking methods based on MLE cannot directly optimize the future  reward nor conduct multi-objective optimization. (2) Q-learning has low variance and is off-policy learning, but it is inherently not suitable for ranking and suffers from the overestimation issue. To address these issues, in this section, we propose a novel off-policy value ranking algorithm. 
\subsection{The One-step EM framework}
\label{Sec:One}
In this section, we first present the EM framework which can unify the MLE and RL models to learn from both the reward and ranking signal on the single-step RL environments (bandit setting). This bandit setting provides the reader with a simplified version of our main contribution in this paper, i.e., the off-policy value ranking algorithm.

To this end, we adopt the principle of probabilistic generative model and  introduce a binary variable $R$ which is related to the reward by $p\left(R_{t}=1 | {s}_{t}, a_{t}\right)=\exp (\frac{r\left({s}_{t}, a_{t}\right)}{\alpha})$ where $\alpha$ is the temperature parameter, to denote whether the action $a_t$ taken  at state $s_{t}$ is optimal. A larger reward means that conducting action $a_t$ at state $s_t$ is more likely to be optimal. We use $R_{t}$ to represent $R_{t}=1$ for brevity.  Without loss of generality, we assume the rewards are nonpositive as the rewards can always be scaled and centered to be no greater than 0, which guarantees that $\exp (\frac{r\left({s}_{t}, a_{t}\right)}{\alpha})\le 1$. To connect MLE with RL optimization, we further consider the model $p_{\theta}(a_t | s_t)$ as a policy that takes action $a_{t}$ given the state $s_{t}$. Given the definitions above, we consider the following  probabilistic generative process of rewards and feedbacks:
\begin{linenomath}
\small
\begin{align}
 \quad  a_t \sim p_{\theta}(a_t | s_t), \quad R_{t} \sim p\left(R_{t} | {s}_{t}, a_{t}\right).
\end{align}
\end{linenomath}
For a given state $s_t$, the recommendation system selects an action $a_t$ based on the optimizing policy and  the optimal variable $\mathcal{R}_{t}$ is generated by the probability $p\left(R_{t} | {s}_{t}, a_{t}\right)$ based on the received reward $r({s}_{t}, a_{t})$. This generative model can be represented  as a joint probability:
\begin{linenomath}
\footnotesize
\begin{align}
p(a_t, R_{t}|s_t)=p_{\theta}(a_t | s_t) p\left(R_{t} | {s}_{t}, a_{t}\right). \label{Bandit}
\end{align}
\end{linenomath}
Note that we let the probability conditioned on $s_t$ since $p({s}_{t})$ is the empirical state distribution which is known given logged feedbacks. To learn this generative model,  we are interested in estimating $\theta$ and we can directly maximize the log-marginal likelihood $\log p_{\theta}(a_{t}|s_t)$ to learn the $\theta$ based on the observed feedbacks $(s_t,a_t)$, resulting in the  probability supervised ranking method with MLE. However, instead of conducting MLE, we want to learn the optimal policy directly from user rewards.  Thus, we 
consider maximizing the log-marginal likelihood $\log p_{\theta}(R_{t}|s_t)$ of observed optimal variable $R_{t}$ and treat the action $a_{t}$ as the latent variable:
\begin{linenomath}
\footnotesize
\begin{align} 
&\log p_{\theta}(R_{t}|s_t)=\log \int p_{\theta}\left(a_t|s_t, R_{t}\right) p_{\theta}(R_{t}|s_t) da_t \nonumber \\
=& \int p_{\theta}\left(a_t|s_t, R_{t}\right)  \log \frac{p_{\theta}\left(a_t, R_{t}|s_t\right)}{p_{\theta}\left(a_t|s_t, R_{t}\right) }da_t \label{Eq:bandit1} \\
=&\mathbb{E}_{p_{\theta}\left(a_t|s_t, R_{t}\right)}[\frac{r(a_t,s_t)}{\alpha}+ \log p_{\theta}(a_t|s_t)-\log p_{\theta}\left(a_t|s_t, R_{t}\right)]. \nonumber 
\end{align}
\end{linenomath}
Eq.~(\ref{Eq:bandit1}) shows that maximizing the log-marginal likelihood $\log p_{\theta}(R_{t}|s_t)$ is equivalent to maximizing the reward at time $t$ with the posterior  policy $p_{\theta}\left(a_t|s_t, R_{t}\right)$, while minimizing the Kullback-Leibler (KL) divergence between $p_{\theta}\left(a_t|s_t, R_{t}\right)$ and $p_{\theta}\left(a_t|s_t\right)$. $\alpha$ serves as a trade-off parameter here. When $\alpha\rightarrow 0$, this objective is equivalent to only maximizing the reward at time $t$.  The posterior $p_{\theta}\left(a_t|s_t, R_{t}\right)$  is closely related to the notion of the optimal policy and this can be intuitively understood as: “What is the probability of action given my current state if we want to be  optimal?''

To answer this question, for each state, we need to learn the posterior distribution $p(a_t| s_t, R_{t})$. We resort to the EM algorithm~\cite{neal1998view}   that iterates two coordinate ascent optimization steps. We assume a variational distribution $q(a_t|s_t)$. 
Given this, we could derive the surrogate function to lower-bound the log-marginal likelihood $\log p_{\theta}(R_{t}|s_t)$ of observed optimal variable $R_{t}$ as follows:
\begin{linenomath}
\small\begin{align}
&\log p_{\theta}(R_{t}|s_t)=\log \int p_{\theta}(a_t,R_{t}|s_t) da_t= \nonumber \\
&\int q_{} (a_t |s_t) \log \frac{p_{\theta}\left(a_t, R_{t}|s_t\right)}{q_{}(a_t|s_t)}da_t +KL\left(q_{{}}\left(a_t |s_t\right)||p_{\theta}\left(a_t|s_t, R_{t}\right) \right) \nonumber \\
&\geq \int q_{} (a_t |s_t) \log \frac{p_{\theta}\left(a_t, R_{t}|s_t\right)}{q_{}(a_t|s_t)}da_t \triangleq \mathcal{L}(q, {\theta}),
\end{align}
\end{linenomath}
where the inequality holds since the KL divergence is always non-negative. 
$\mathcal{L}(q, {\theta})$ is known as Evidence Lower BOund (ELBO)~\cite{blei2017variational,xiao2021learning}. Instead of directly maximizing the  log-likelihood, the EM algorithm maximizes  ELBO $\mathcal{L}(q, {\theta})$ via an alternative procedure: \\
\textbf{E-step}. At $n$-th iteration, given the current  $\theta_{n}$, the E-step that maximizes $\mathcal{L}(q, {\theta})$ w.r.t $q$ has a closed-form solution: 
\begin{linenomath}
\footnotesize
\begin{align}
q_{n+1}(a_t| s_t)=p_{\theta_n}(a_t|s_t, R_{t}) \propto p_{\theta_n}(a_t| s_t) \exp (\frac{ r(s_t, a_t )}{\alpha}). \label{Eq:E}
\end{align}
\end{linenomath}
\textbf{M-step}. Given  $q_{n+1}(a_t| s_t)$ at last E-step, we maximize $\mathcal{L}(q, {\theta})$ w.r.t $\theta$, resulting in the following  objective:
\begin{linenomath}
\small
\begin{align}
 {\theta}^{n+1}=\operatorname{argmax}_{\theta} \mathbb{E}_{q_{n+1}(a_t| s_t)}\left[\log p_{\theta}(a_t| s_t)\right]. \label{Eq:M}
\end{align}
\end{linenomath}
Since there is no analytical solution for $\theta$ which is typically a neural network, we update $\theta$ via stochastic gradient descent with samples from $q_{n+1}(a_t| s_t)$ at the M-step in general.

In E-step, $q(a_t| s_t)$ is constructed by projecting $p(a_t| s_t)$ into a subspace constrained by the rewards with trade-off $\alpha$ in Eq. (\ref{Eq:E}), and thus has desirable properties in learning rewards. In M-step, $q(a_t| s_t)$ serves as a teacher to transfer this reward knowledge into the  student policy $p(a_t| s_t)$.  This formulation can be understood as a type of supervised knowledge distillation~\cite{hinton2015distilling} and unsupervised posterior regularization~\cite{ganchev2010posterior}. 

However, \textbf{fundamentally different from  supervised and unsupervised learning, it is non-trivial to conduct the EM algorithm in the recommendation setting.} (1) Partial and sparse rewards. There is no way to directly access the reward for every state-action pair. In other words, we only observe reward $r_{t}$  from logged feedbacks $\mathcal{D}$, but not the others. This limits us to conduct the E-step since it requires the reward  $r(a_t, s_{t})$ for every state-action pairs.  In addition, the reward is relatively sparse and lacks a negative signal. (2) Future rewards. Although this formulation can maximize the reward at the time $t$,  it  can not maximize long-term rewards. Before we discuss how to maximize  long-term rewards, we first propose  reward extrapolation and ranking regularization to solve the  challenge of partial and sparse rewards.

\subsection{Reward Extrapolation and Ranking Regularization}
\label{sec:ER}
\textbf{Extrapolation for partial rewards in the E-step}. At each iteration $n$, we propose to extrapolate the reward from the logged feedbacks.  Instead of extrapolating with worst-case rewards, i.e., setting the reward to zero for every unseen state-action pair, we can use a reward regressor  to  reduce the estimation  bias. In particular, we extrapolate  reward using the following regression estimator over mini-batches of tuples $\left({s}_{t}, a_{t}, r_t, {s}_{t+1}\right) \in \mathcal{D}$ with the MSE loss:
\begin{linenomath}
\footnotesize
\begin{align}
\mathcal{L}_{\text{Q}}(\phi)=(Q_{\phi}\left(s_{t}, a_{t}\right)-r_t)^{2}, \label{Eq:Re}
\end{align}
\end{linenomath} 
where the reward estimator $Q_{\phi}\left(s_{t}, a_{t}\right)$ can be arbitrary models such as neural networks. Generally,  it can  extrapolate well since the  error is relatively low due to the expressivity of the function approximator. However, if the logging policy is poor and differs from the optimizing policy $p_{\theta}(a_t|s_t)$, minimizing Eq.(\ref{Eq:Re}) would make the  regression model fits well for items that are far from optimal, resulting in biased model. 
To alleviate this, we adopt importance sampling to shift the regression model to the optimizing policy:
\begin{linenomath}
\footnotesize
\begin{align}
{\mathcal{L}}_{\text{Q}}(\phi)=\frac{q_{n}(a_t|s_t)}{{p}_{\psi}(a_t|s_t)}(Q_{\phi}\left(s_{t}, a_{t}\right)-r_t)^{2}, \label{Eq:Re2}
\end{align}
\end{linenomath}
where ${p}_{\psi}\left(a_{t} | s_{t}\right)$ is the estimated logging policy from logged feedbacks via simple MLE estimation in Eq.~(\ref{MLE}). This  objective corrects the selection bias induced by the logging policy  so that it becomes in expectation equivalent to training the estimator with on-policy data from $q_{n}(a_t|s_t)$. Although the huge action space could lead to very small importance weights for some items, and consequently greater variance~\cite{dudik2011doubly}, in \S~\ref{sec:Discussion}, we theoretically show that our EM framework can naturally reduce the  variance of this estimator. Given  $Q_{\phi}\left(s_t, a_t \right)$ and Eq.~(\ref{Eq:E}), the posterior in the E-step can be derived as follows:
\begin{linenomath}
\footnotesize
\begin{align}
q_{n+1}(a_t| s_t) \propto p_{\theta_{n}}(a_t| s_t) \exp \left(\frac{ Q_{\phi}\left(s_t, a_t \right)}{\alpha}\right).
\end{align}
\end{linenomath}
\textbf{Regularization for sparse rewards in the M-step}. The M-step, i.e., Eq.~(\ref{Eq:M}), is for optimizing the policy parameter $\theta$. Although the teacher $q_{n+1}(a_t| s_t)$ can transfer the reward knowledge into $p_{\theta}(a_t|s_t)$, the reward is relative sparse and lacks negative signal.  To make the policy $p_{\theta}(a_t | s_t)$  be able to take into account negative evidence, we modify the M-step in Eq.~(\ref{Eq:M})  by adding a ranking regularization term as:
\begin{linenomath}
\footnotesize
\begin{align}
\mathcal{L}&_{\text{P}}({\theta})=\underbrace{\beta\mathbb{E}_{q} [\log p_{\theta}(a_t| s_t)]}_{\text{Reward}}+
(1-\beta)\underbrace{\mathbb{E}_{{p}_{\psi}} [\log p_{\theta}(a_{t} | s_{t})] }_{\text{Ranking}}, \label{Eq:MM}
\end{align}
\end{linenomath}
where  $q \triangleq q_{n+1}(a_t| s_t)$ and $\beta$ is the trade-off parameter calibrating the relative importance of the two objectives. ${p}_{\psi} \triangleq {p}_{\psi}\left(a_{t} | s_{t}\right)$ is the estimated ranking policy from logged feedbacks via simple MLE estimation (list-wise) and serves as regularization for RL policy $p_{\theta}(a_t|s_t)$. The key difference between reward regression $q$ and ranking regularization ${p}_{\psi}$ is that ranking regularization must produce a normalized distribution over actions which does  constrain the probability of non-clicked recommendations (unobserved actions). In contrast,
reward regression needs to estimate the future rewards  of each state-action pair  which can not be normalized. 
By alternately conducting the modified E-step in Eq.~(\ref{Eq:Re2}) and M-step in Eq.~(\ref{Eq:MM}),  $\alpha$ and $\beta$ provide trading-off between the reward and  signal of the ranking metric in the optimizing  process of the teacher $q(a_t|s_t)$ and student $p_{\theta}(a_t|s_t)$. 


%
\subsection{The Sequential EM Framework}
\label{sec:SL}
In the last section, we have introduced how to learn a ranking algorithm from both rewards and ranking signals. However, the bandit setting does not incorporate any information about the dynamics, so it learns to greedily assign high values to  actions and  can not maximize  future rewards, without considering the state transitions that occur as a consequence of actions. In other words, in the full RL setting, the state distribution $p(s_t)$ is not independent and identically distributed (i.i.d) now but depends on the action caused by the policy in the last step, i.e., the transition probability. 
We extended the one-step EM  framework for the bandit setting in \S~\ref{Sec:One} into the sequential setting. Similar to what we did in  \S~\ref{Sec:One}, we 
introduce a binary variable $R_t$: $p\left(R_{t}=1 | {s}_{t}, a_{t}\right)=\exp (\frac{r\left({s}_{t}, a_{t}\right)}{\alpha})$  to denote whether the action $a_t$ taken at state $s_{t}$ is optimal.
With the  definition of MDP, we have the following generative model with the  trajectory $\tau=\left\{{s}_{1}, a_{1}, \cdots, {s}_{T}\right\}$ and optimal variables $R_{{1:T}}$:
\begin{equation}
\small
p_{\theta}(\tau, R_{1:T})=\rho({s}_{1}) \prod_{t=1}^{T} p({s}_{t+1} | {s}_{t}, a_{t}) p(R_{t} | {s}_{t}, a_{t}) p_{\theta}(a_{t}| s_t ).
\label{Eq:Full}
\end{equation}
Compared to  the bandit setting in Eq.~(\ref{Bandit}), the distribution of state is non-i.i.d  now but depends on the last step action.  Given this generative model, we can also conduct the MLE estimation over the logged trajectory, resulting in Eq.~(\ref{MLE}). However, MLE can not learn from rewards. In addition, the MLE separates the logged feedbacks in the trajectory which makes it unable to maximize the long-term reward. Thus, like the bandit setting in Eq.~(\ref{Bandit}), we are interested in maximizing the log-marginal likelihood $\log p_{\theta}(R_{1:T} |\tau)$ the inferring the posterior $p(\tau | R_{1:T})$.
In particular, the posterior $p(\tau | R_{1:T} )$ can be factorized as follows due to the conditional dependency underlaying the generative model in Eq. (\ref{Eq:Full}).
\begin{linenomath}
\small
\begin{align}
p(\tau|R_{1:T})=\prod_{t=1}^{T}p({s}_{t}|{s}_{t-1}, {a}_{t-1},R_{t:T})p(a_t|{s}_{t},R_{t:T}),
\end{align}
\end{linenomath}
where  $p({s}_{1}|{s}_{0}, {a}_{0},R_{1:T})=p({s}_{1}|R_{1:T})$. Following the derivation of Eq.~(\ref{Eq:bandit1}), one can easily verify that maximizing the log-marginal likelihood $\log p_{\theta}(R_{1:T} |\tau)$ is equivalent to maximizing our RL goal Eq.~(\ref{Eq:RL1}) with optimal policy $p(a_t|{s}_{t},R_{t:T})$. Thus, the probability of action given optimality of the current until end of the episode and current state, i.e., the  posterior of action $p(a_t | {s}_{t},R_{t:T})$  is of high interest.  Unfortunately, It is  difficult to get the posterior $p(a_t| {s}_{t},R_{t:T})$ since the distribution of state is non-i.i.d. 
\subsection{Off-Policy Value Ranking}
\label{sec:OPVR}
In this section, we propose an EM-style algorithm to approximate the posterior, which results in our off-policy  value ranking algorithm.  Recall that our goal is  to infer the posterior $p(a_t | {s}_{t},R_{t:T})$. We consider inferring  the posterior through the  message passing algorithm~\cite{heskes2002expectation}. Thus, we  first define the backward messages as:
\begin{linenomath}
\small
\begin{align} 
m\left({s}_{t}, a_{t}\right) \triangleq p\left(R_{t: T} | {s}_{t}, a_{t}\right) && m\left({s}_{t}\right) \triangleq p\left(R_{t: T} | {s}_{t}\right). \label{Eq:m2}
\end{align}
\end{linenomath}
Then our interested posterior  $p(a_t | {s}_{t},R_{t:T})$ is:
\begin{linenomath}
\small
\begin{align} 
&p(a_{t} | {s}_{t}, R_{t: T})=\frac{p({s}_{t}, a_{t} | R_{t: T})}{p({s}_{t} | R_{t: T})} =\frac{p(R_{t: T} | {s}_{t}, a_{t}) p_{\theta}(a_{t} | {s}_{t}) p({s}_{t})}{p(R_{t: T} | {s}_{t}) p({s}_{t})}\nonumber \\
=& 
\frac{m({s}_{t}, a_{t})}{m({s}_{t})} p_{\theta}(a_t |s_t)\triangleq q(a_t|s_t). \label{Eq:policy}
\end{align}
\end{linenomath}
With Eq. (\ref{Eq:Full}), reward extrapolation, and 
ranking regularization proposed in \S~\ref{sec:ER}, we can obtain the following E and M-steps (see appendix in SM for  detailed derivations):\\
\textbf{E-step}. At  the $n$-th iteration, we minimize the loss w.r.t $\phi$:
\begin{linenomath}
\small
\begin{align}
\mathcal{L}_{\text{Q}}({\phi})&=\frac{q_{n}(a_t|s_t)}{{p}_{\psi}(a_t|s_t)} \big(Q_{{\phi}}\big({s}_{t}, a_{t}\big)-r_t -\gamma \big(Q_{\bar{{\phi}}}\left({s}_{t+1}, a_{t+1}\right) \nonumber \\
+&\alpha \log { p_{\theta_{n}}(a_{t+1}|{s}_{t+1})} -\alpha \log{  q_{n}\left(a_{t+1} | {s}_{t+1}\right)}\big )\big)^{2}, \label{Eq:Q}
\end{align}
\end{linenomath}
where the $\bar{\phi}$ indicates parameters of the target network  the same as vanilla Q-learning in Eq.~(\ref{TD}) and $a_{t+1}$ is sampled from the policy $ q_{n}\left(a_{t+1} | {s}_{t+1}\right)$.  We add the importance sampling as same as we did in the bandit setting in Eq.~(\ref{Eq:Re2}). We can find the reward extrapolation Eq.~(\ref{Eq:Re2}) in the batch setting becomes the Q value estimation for estimating future rewards.  Given Eq.~(\ref{Eq:policy}), we can get the followings non-parametric  optimal policy  $p(a_t|{s}_{t},R_{t:T})$:
\begin{linenomath}
\small
\begin{align}
q_{n+1}(a_t| s_t)\propto p_{\theta_n}(a_t| s_t) \exp \left(\frac{ Q_{\phi}\left(s_t, a_t \right)}{\alpha}\right).\label{Eq:Q2}
\end{align}
\end{linenomath}
Comparing with the E-step (Eq.~(\ref{Eq:Re})), the reward estimator $Q_{\phi}(s_t, a_t)$ becomes the value function approximating future rewards. When discount factor $\gamma=0$, the objective Eq.~(\ref{Eq:Q}) is analogous  to  the Eq.~(\ref{Eq:Re}) in the bandit setting. \\
\textbf{M-step with ranking regularization}. The M-step is analogous  to  Eq.~(\ref{Eq:MM}) in the bandit setting, and we also consider the following  loss with the ranking  regularization term:
\begin{linenomath}
\small
\begin{align}
\mathcal{L}&_{\text{P}}({\theta})=\beta{\mathbb{E}_{q_{}}[\log p_{\theta}(a_t| s_t)]}+
(1-\beta){\mathbb{E}_{p_{\psi}}[\log p_{\theta}(a_{t} | s_{t})]},
\label{Eq:VMM}
\end{align}
\end{linenomath}
where $q\triangleq q_{n+1}(a_t| s_t)$. By alternately conducting the E-step in Eq.~(\ref{Eq:Q2}) and M-step in Eq.~(\ref{Eq:VMM}), we get our full VR algorithm which can simultaneously maximize the  future long-term rewards and learn from the ranking signal.

\begin{table*}[!t]
\center
\caption{ Overall performance comparison.  The percentage in brackets denote the relative performance improvement over
MLE.}
\label{Table:Overall}
\vskip -1em
\resizebox{1.0\textwidth}{!}{
\begin{tabular}{@{}l ccccc cc cccccc c  @{}}
\toprule[1.0pt]
 \multirow{2}{*}{Backbones} &\multirow{2}{*}{Methods}& \multicolumn{4}{c}{YooChoose}  & & \multicolumn{4}{c}{RetailRocket}   \\

  \cline{3-6}    \cline{8-11} 
 & & HR$@$5  & NDCG$@$5     &  HR$@$20  & NDCG$@$20   &   &  HR$@$5  & NDCG$@$5  &HR$@$20  & NDCG$@$20 &  \\
 
       \toprule[1.0pt]
    \multirow{3}{*}{SASRec} &   MLE  &  0.2811  & 0.1857  &  0.4323  & 0.2389   & & 0.2109  & 0.1512   &   0.3136  & 0.1918  &  \\
    &  PG &  0.2968(5.6\%)  & 0.1901(2.4\%)   &  0.4417(2.1\%)  & 0.2511(5.1\%)   & &   0.2235(6.0\%)  & 0.1638(8.1\%)   &    0.3211(2.4\%)   & 0.2102(9.6\%)&  \\
    &VR  &  \textbf{0.3187}(13.4\%) & \textbf{0.2033}(9.5\%)   &  $\textbf{0.4618}$(6.8\%)   &  $\textbf{0.2635}^{}$(10.3\%)    & & \textbf{0.2326}(10.3\%) & \textbf{0.1729}(14.4\%)    &  $\textbf{0.3429}^{}$(9.3\%)  & $\textbf{0.2336}^{}$(21.8\%) &  \\
    \hline
    \multirow{3}{*}{GRU4Rec}    &MLE  &  0.2614  & 0.1599 &  0.4238  & 0.2117   & & 0.1947  & 0.1431   &  0.2987 & 0.1856  &  \\
   &PG  &  0.2773(6.1\%)   & 0.1726(7.9\%) &  0.4391(3.6\%)  & 0.2236(5.6\%)  & & 0.2068(6.2\%)  & 0.1543(7.8\%)    &  0.3152(5.5\%)   & 0.1991(7.3\%)  &  \\
    &VR  &  \textbf{0.2905}(11.1\%)  & \textbf{0.1887}(18.0\%)   &  \textbf{0.4501}(6.2\%)  & $\textbf{0.2457}(16.1\%)$  & & \textbf{0.2188}(12.4\%) & \textbf{0.1657}(15.8\%)     &  $\textbf{0.3372}^{}$(12.9\%)  & $\textbf{0.2208}^{}$(19.0\%)   &  \\
    \hline
   \multirow{3}{*}{Caser}    &MLE  &  0.2521  & 0.1585  &  0.4171  & 0.2016   & &0.1776  & 0.1255   &   0.2845  & 0.1739  &  \\
   &PG   &  0.2655(5.3\%)  & 0.1677(5.8\%)  &  0.4283(2.7\%)  & 0.2132(5.8\%)     & &  0.1901(7.0\%)   & 0.1387(10.5\%)    & 0.2989(5.1\%) & 0.1822(4.8\%)  &  \\
   &VR  &  \textbf{0.2809}(11.4\%)   & \textbf{0.1759}(11.0\%)  & $ \textbf{0.4426}^{}$(6.1\%)  & $\textbf{0.2371}^{}$(17.6\%)   & & \textbf{0.2075}(16.8\%)   & \textbf{0.1451}(15.6\%)   &   $\textbf{0.3205}^{}(12.7\%)   $ & $\textbf{0.2119}^{}$(21.9\%)  &  \\
    \bottomrule[1.0pt]
\end{tabular}}
\vskip -1.5em
\end{table*}

\section{Theoretical Analysis}
\label{sec:Discussion}
\textbf{Overestimation Bias}:
\label{subsec:OB}In this section, we show that our VR can effectively reduce the overestimation bias compared with Q-learning. Recall that the overestimation bias occurs due to the max operator $\max _{a} Q_{\bar{\phi}}\left({s}_{t+1}, a\right)$ in the TD loss of Q-learning. Specifically, the overestimation bias at state $s$ can be represented as:
$\max _{a}Q_{\bar{\phi}}(s, a)-\max _{a}Q(s, a) = \max _{a}(Q_{\bar{\phi}}(s, a)-V(s)) 
=\max_{a} \left(\epsilon_{a}\right)$,
where $V(s)$ and $Q(s, a)$ are the true state and state-action value functions, respectively. Since our TD loss in Eq.~(\ref{Eq:Q}) selects the action from the  policy $q(a|s)$ in the E-step, the overestimation bias  can be proved to be no greater than that in Q-learning:
\begin{linenomath}
\scriptsize
\begin{align}
& \sum_{a} q(a| s) Q_{\bar{\phi}}(s, a)-V(s)= \sum_{a} \frac{p_{\theta}(a'| s) \exp (\frac{ Q_{\bar{\phi}}(s, a')}{\alpha})}{\sum_{{a'}}p_{\theta}(a'| s) \exp (\frac{ Q_{\bar{\phi}}(s, a' )}{\alpha})}Q_{\bar{\phi}}(s, a) \nonumber \\
&-V(s)=\sum_{a} \frac{p_{\theta}(a'| s) \exp (\frac{ V(s)+\epsilon_{a'}}{\alpha})}{\sum_{{a'}}p_{\theta}(a' | s) \exp (\frac{V(s)+\epsilon_{a'}}{\alpha})}(V(s)+\epsilon_{a})-V(s) \nonumber \\
&=\sum_{a} \frac{p_{\theta}(a'| s)\exp  \left[\frac{ \epsilon_{a'}}{\alpha}\right]}{\sum_{{a'}} p_{\theta}(a'| s) \exp \left[\frac{ \epsilon_{{a'}}}{\alpha}\right]} \epsilon_{a} \leq  \max _{a}\left(\epsilon_{a}\right).
\end{align}
\end{linenomath}
\textbf{Estimation Variance}:
\label{subsec:EV}
We theoretically show that our M-step can bound the variance induced by the importance sampling in the unbiased reward extrapolation. For notation brevity, we denote unbiased reward estimators  (Eqs.~(\ref{Eq:Re2}) and (\ref{Eq:Q})) as $w({a}_{t}) \mathcal{L}_{\text{Q}}\triangleq\mathcal{L}_{\text{Q}}^{w}$ in which $w({a}_{t})=\frac{q_{n}(a_t|s_t)}{{p}_{\psi}(a_t|s_t)}\triangleq \frac{q}{{p}}$. we can conclude that the variance of estimation error can be bounded as follows (a proof is provided in the appendix):
\begin{linenomath}
\footnotesize
\begin{align}
&\operatorname{Var}_{{a}_{t} \sim p}[\mathcal{L}_{\text{Q}}^{w}] =\mathrm{E}_{{a}_t \sim p}[(\mathcal{L}_{\text{Q}}^{w} )^{2}]-(\mathrm{E}_{{a}_{t} \sim p}[\mathcal{L}_{\text{Q}}^{w} ])^{2} \leq \\
&d_{\lambda+1}(q \| p)(\mathrm{E}_{{a}_{t} \sim p} \mathcal{L}_{\text{Q}}^{w})^{1-\frac{1}{\lambda}}(\sum_{{a}_{t}} \mathcal{L}_{\text{Q}})^{1+\frac{1}{\lambda}}-(\mathrm{E}_{{a}_{t} \sim p}[\mathcal{L}_{\text{Q}}^{w}])^{2}, \nonumber ~\forall \lambda \geq 0
\end{align}
\end{linenomath}
where $d_{\lambda}(q \| p)=2^{D_{\lambda}(q \| p)}$ and $D_{\lambda}(q \| p)$ is the Rényi divergence~\cite{renyi1961measures}.  Since  $\lim_{\lambda \rightarrow 0}D_{\lambda+1}(q \| p)=KL(q\|p)$ and our M-step can regularize the  policy $q$ towards the logging policy $p_{\psi}$, our VR can  reduce the variance.

%% file: Experiment.tex
\section{Empirical Evaluation}
In this section, we empirically evaluate the effectiveness of the proposed VR on both offline and online settings.
\subsection{Offline Ranking Experiment}
\label{Overall performance}
\textbf{Datasets}. We use two large-scale  datasets.
(1) \textit{YooChoose}\footnote{ \url{https://recsys.acm.org/recsys15/challenge/}.}: It contains sequences of user purchases and clicks. We remove the sessions whose length is smaller than 3. (2) \textit{RetailRocket}\footnote{\url{https://www.kaggle.com/retailrocket/ecommerce-dataset}}: This dataset also contains sequences of user purchases and clicks. We remove  items which are interacted less than 3 times. We randomly sample 80\% sequences as the training set, 10\% as validation, and 10\% as the test set. \\
\textbf{Settings}. Our VR is a generic learning algorithm and can be utilized to improve any recommendation backbones. To evaluate the effectiveness of VR, we consider three representative  recommendation backbones, i.e.,  GRU4Rec~\cite{hidasi2016session}, Caser~\cite{tang2018personalized} and SASRec~\cite{kang2018self}. Following ~\cite{DBLP:conf/wsdm/ChenBCJBC19},  we reuse the user state representation generated from the backbone models, and model SL policy $p_{\psi}$, Q function $Q_{\phi}$, and RL policy $p_{\theta}$ with different heads. We compare with two learning baselines: the supervised learning with MLE, and the off-policy policy gradient (PG) with the weight capping~\cite{DBLP:conf/wsdm/ChenBCJBC19}. Since 
we have compared with Q-learning in \S~\ref{sec:RLTR}, we omit it in this section due to space limitations. We use top-k Hit Ratio (HR$@$k) and  Normalized Discounted Cumulative Gain (NDCG$@$k) to evaluate the ranking performance of the click feedback.\\
\textbf{Results}. Table~\ref{Table:Overall} summarizes the ranking performance of different learning algorithms with three backbones. From Table~\ref{Table:Overall}, we have the following  findings: (1) Our VR consistently and significantly outperforms the PG. This is because our framework has the advantage of simultaneously maximizing future rewards and leveraging the signal of ranking metric from logged feedbacks. This validates the effectiveness of  VR and our motivation that feedback information is important for better ranking. (2)  VR outperforms MLE in all settings, which shows that our VR as an off-policy RL algorithm can outperform supervised MLE via maximizing the future reward although there is no online interaction. (3)  VR significantly outperforms the MLE and PG with all backbones on two datasets, which demonstrates the effectiveness of VR and also shows that our VR  is robust to different backbones and datasets. (4) We observe that our VR achieves better performance on sizes 5 and 20 of ranking list k, which suggests that our VR is agnostic to the size of top-k list.

\begin{figure}[t]
\centering
  \subfigure{
    \label{subfig:SCNDCG.pdf}
    \includegraphics[width=0.145\textwidth]{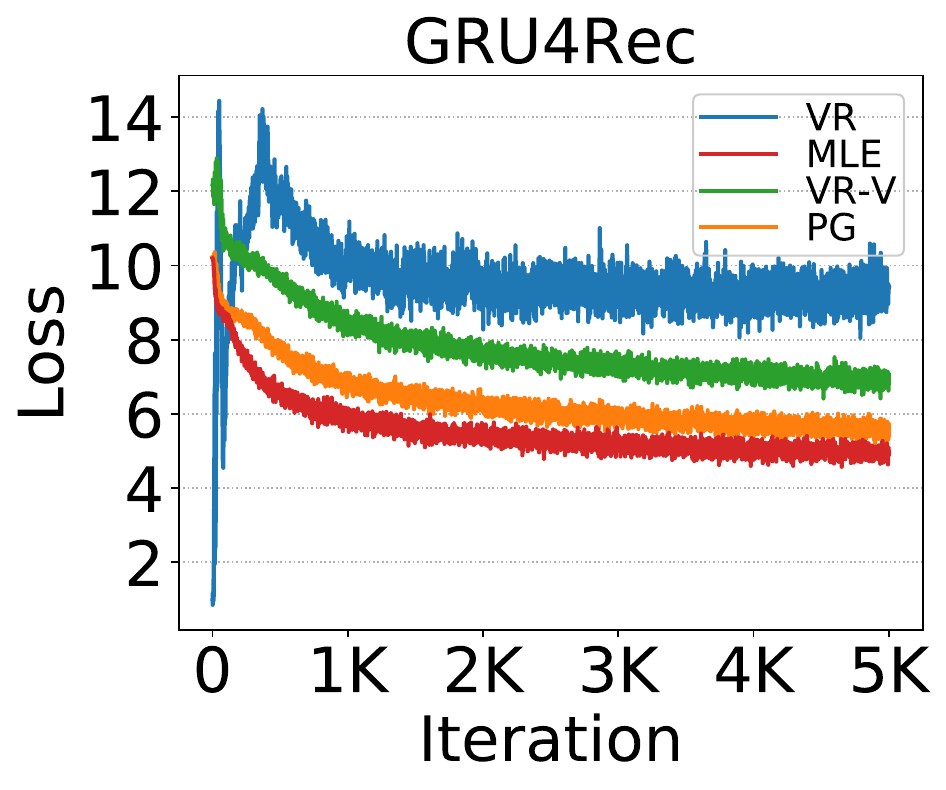}}
      \subfigure{
    \label{subfig:DCNDCG}
    \includegraphics[width=0.15\textwidth]{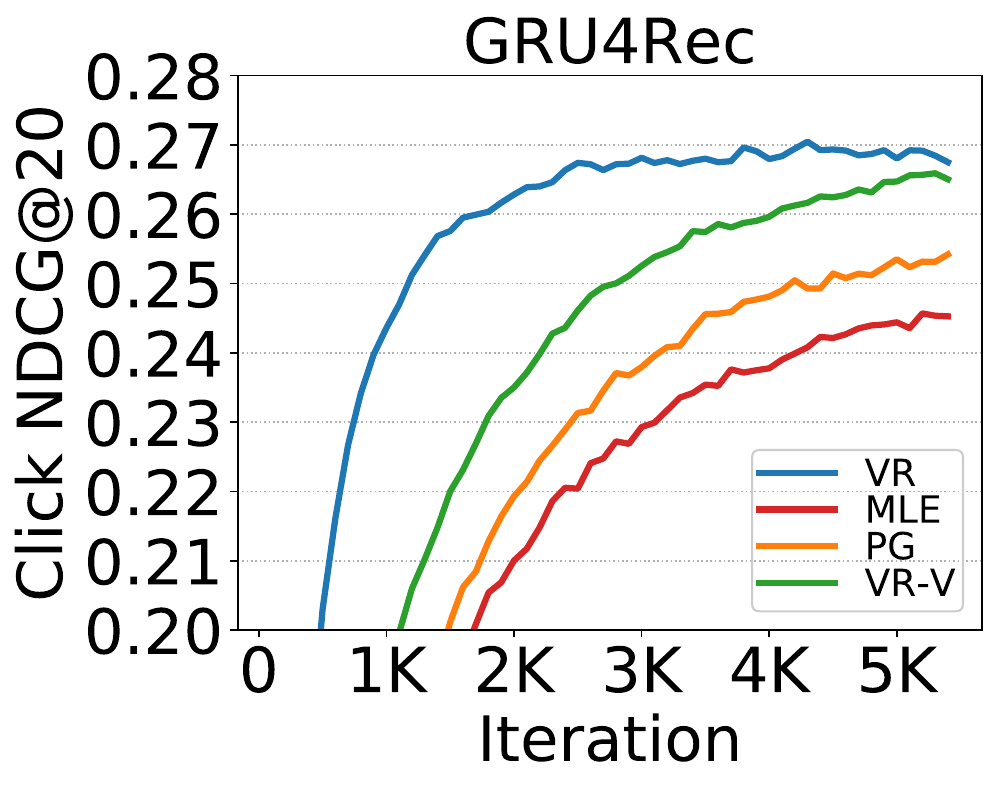}}
       \subfigure{
    \label{subfig:RANDCG}
    \includegraphics[width=0.15\textwidth]{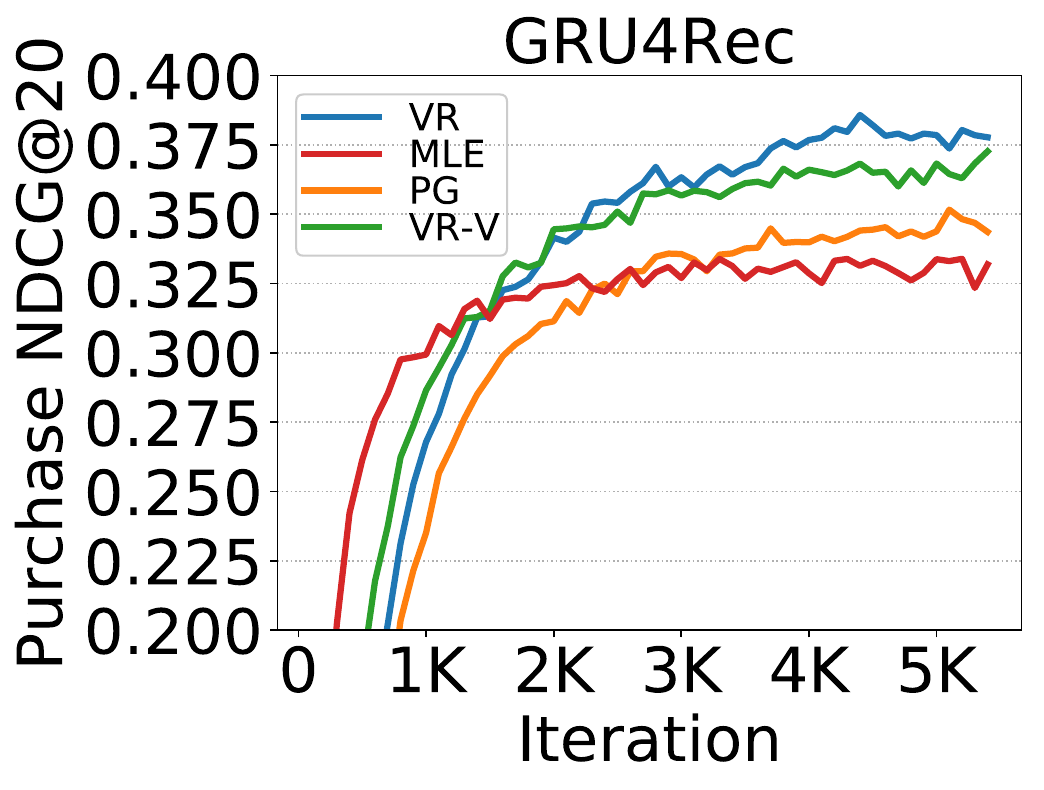}}
         \vskip -0.5em
      \subfigure{
    \label{subfig:SCNDCG.pdf}
    \includegraphics[width=0.145\textwidth]{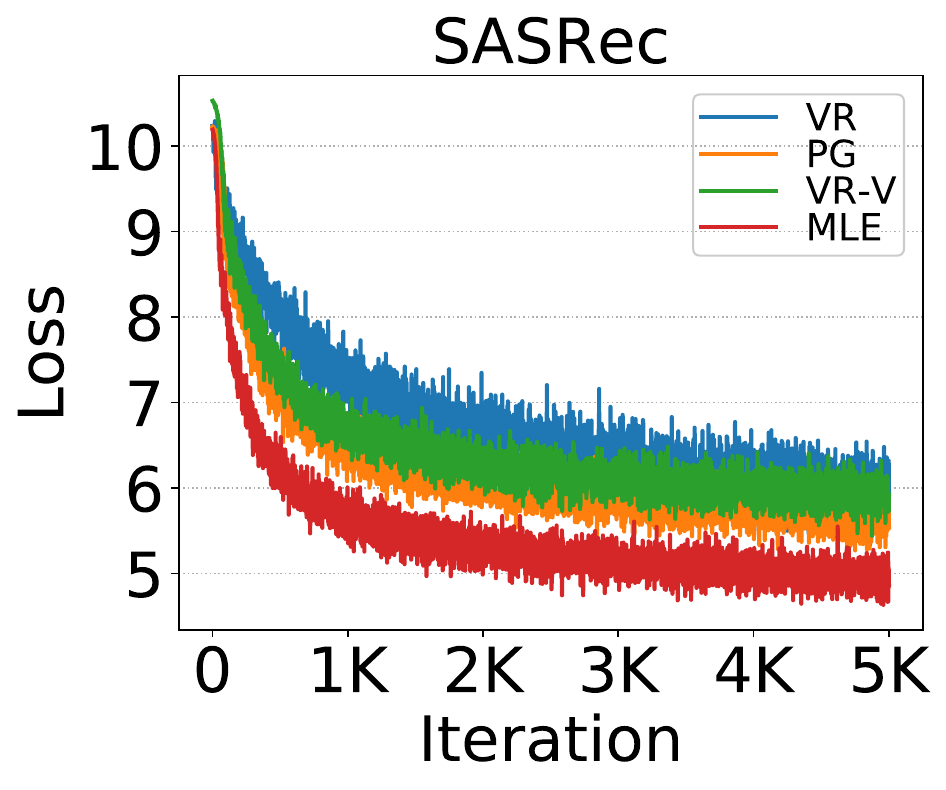}}
      \subfigure{
    \label{subfig:DCNDCG}
    \includegraphics[width=0.15\textwidth]{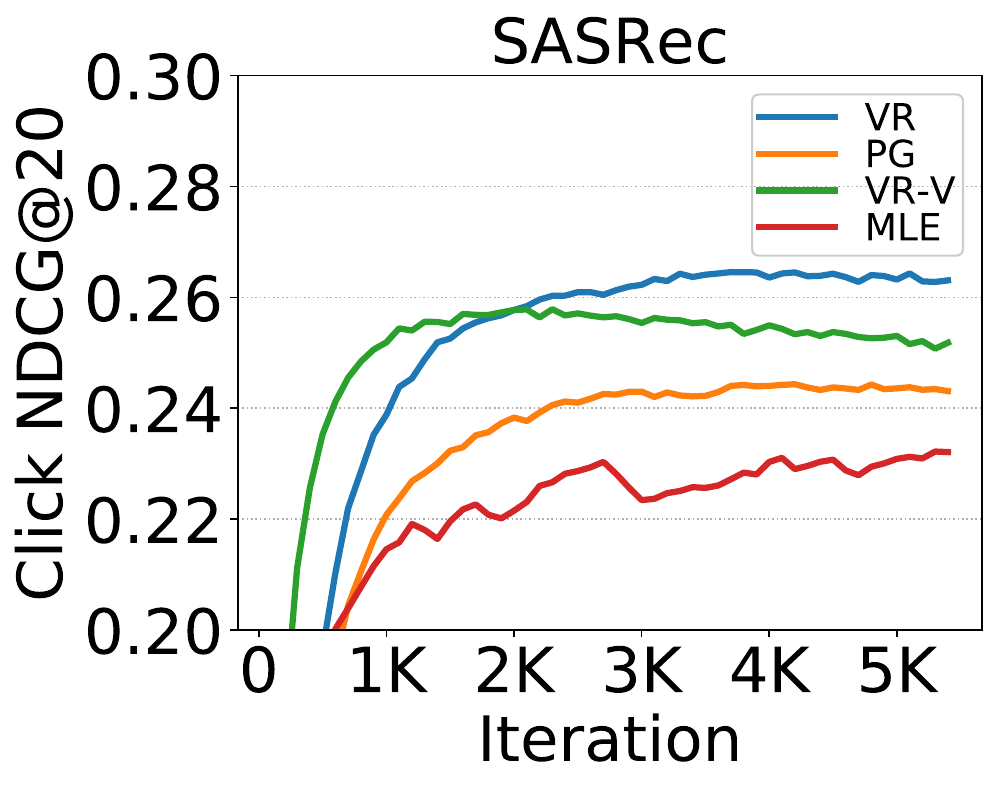}}
      \subfigure{
    \label{subfig:RANDCG}
    \includegraphics[width=0.15\textwidth]{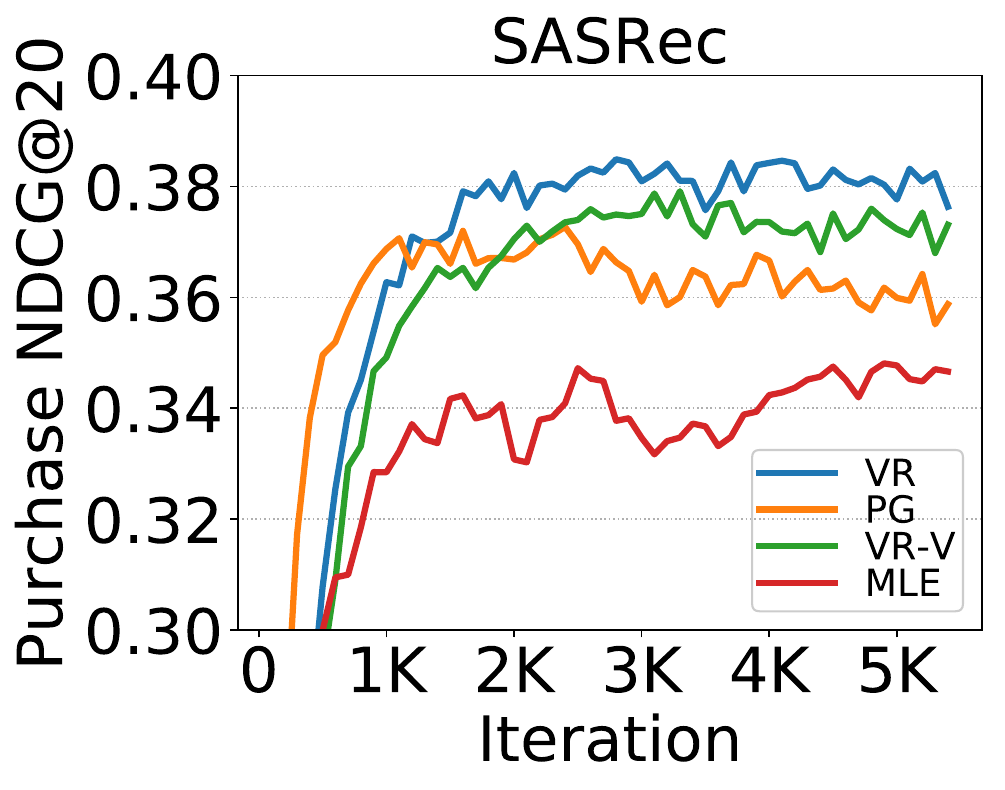}}
    \vskip -1.5em
    \caption{Training curves on the multi-objective setting.}\label{fig:sen}
    \vskip -0.5em
\end{figure} 

\begin{table}[]
\caption{Training time (ms) each iteration.}\label{running_time}
\vskip -1em
\centering
\resizebox{0.35\textwidth}{!}{
\begin{tabular}{lcccccccl}
   \toprule[1.0pt]
\textbf{Backbone}  & \textbf{MLE}    & \textbf{PG} & \textbf{VR-V} & \textbf{VR}  \\
   \toprule[1.0pt]
GRU4Rec & 28.2 & 31.4 & 37.9  & 38.8    \\
Caser & 9.8 & 12.3 & 15.6  & 16.5  \\
SASRec  & 23.1  & 25.2 & 31.6  & 32.5    \\
\toprule[1.0pt]
\end{tabular}}
\vskip -1em
\end{table}

\subsection{Multi-objective Optimization}
\label{exp:RQ2}
One advantage of directly optimizing the reward via RL is that it allows  designing  different rewards  for better optimizing the multi-objectives. In this section, we study how the proposed VR performs in the multi-objective setting. \\
\textbf{Settings}. We follow the same setting in RQ1. However, besides the click, we also consider another feedback, the purchase here. YooChoose contains 43,946 purchases of users. RetailRocket contains 57,269 purchase feedbacks. For VR and off-policy PG, we define the rewards of purchase and click as five and one, respectively. To better show the advantage of VR in the multi-objective setting, we add a variant of our VR called VR-V, which treats the purchase and click the same  and assign the same rewards (one) to them. \\
\textbf{Results}. Fig.~\ref{fig:sen} shows the curves of the policy training loss and testing performance  of purchases and clicks on YooChoose. The results on Caser are similar, so we remove it to Appendix  to save space. According to  Fig.~\ref{fig:sen}, we  find that: (1) Our VR and VR-V outperform  MLE and off-policy PG in all settings, which shows the effectiveness of our method in the multi-objective setting. (2) The phenomenon that VR outperforms VR-V shows that we can achieve better performance by assigning different rewards to different feedbacks. (3) Our VR is training-stable and converges faster than MLE which needs more epochs to achieve stable performance. One possible reason could be that the ranking regularization can reduce the instability in the TD updates. (4) We  find that although  VR has a bigger training loss, it achieves better performance on the test metric which indicates our VR has a much stronger generalization ability. Table~\ref{running_time} shows the results of training time on YooChoose. With Table~\ref{running_time} and Fig.~\ref{fig:sen}, we can find VR is efficient and can achieve superior performance  without much additional computational cost.



\begin{figure}
\vskip -1em
\centering
  \subfigure{
    \includegraphics[width=0.22\textwidth]{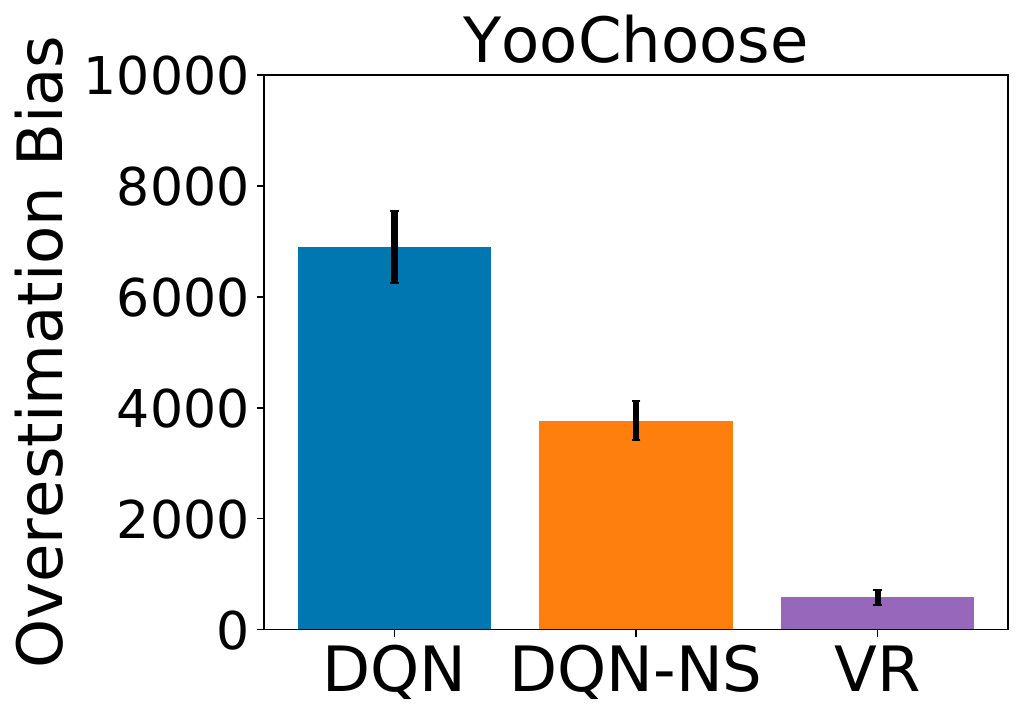}}
  \subfigure{
    \includegraphics[width=0.22\textwidth]{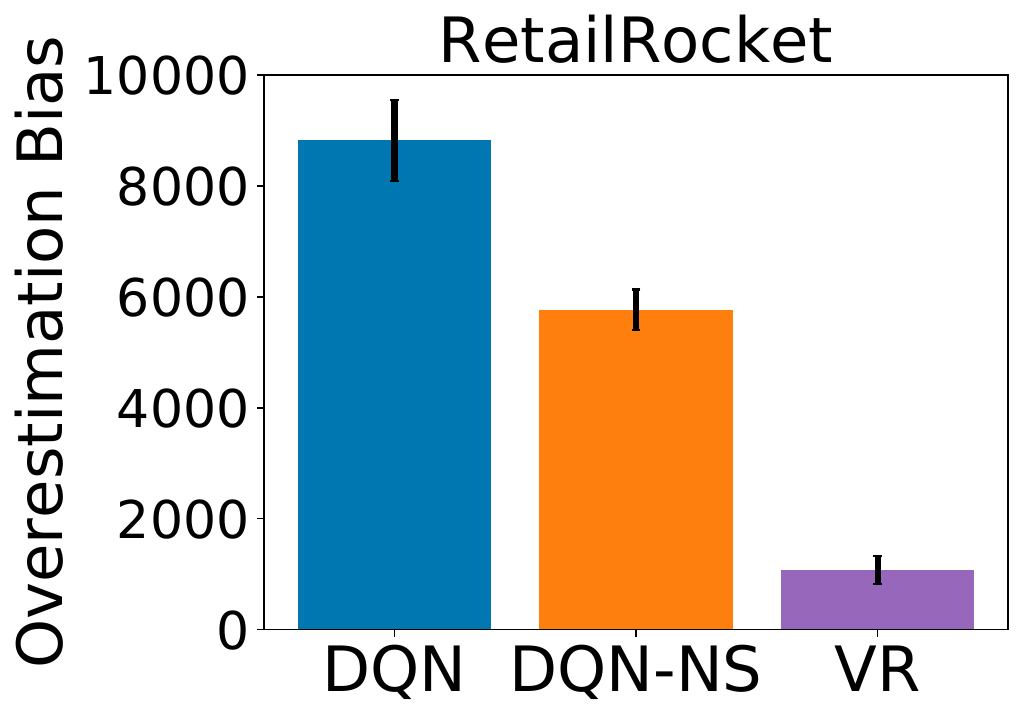}}
    \vskip -1.5em
    \caption{The overestimation bias on two datasets.}\label{fig:Overestimation}
\vskip -0.5em
\end{figure}

\begin{table}[t]
\caption{Ablation study results (\% NDCG@5) varying $\beta$.}\label{Tbl:variance}
\vskip -1em
\centering
\resizebox{0.47\textwidth}{!}{
\begin{tabular}{lccccccl}
   \toprule[1.0pt]
\textbf{Ablation}  & $\beta$=0.2 &  $\beta$=0.4  & $\beta$=0.6 & $\beta$=0.8 \\
   \toprule[0.7pt]
VR-NW &19.21$\pm$0.6 & 19.52$\pm$0.7 & 18.02$\pm$0.5 & 17.67$\pm$0.6   \\ 
VR & \textbf{19.92}$\pm$ 0.9 & \textbf{20.43}$\pm$1.1    & \textbf{19.56}$\pm$1.5 & \textbf{18.92}$\pm$1.8 \\
\toprule[1.0pt]
\end{tabular}}
\vskip -1em
\end{table}

 \begin{figure}[!htb]
\centering
  \subfigure{
    \includegraphics[width=0.22\textwidth]{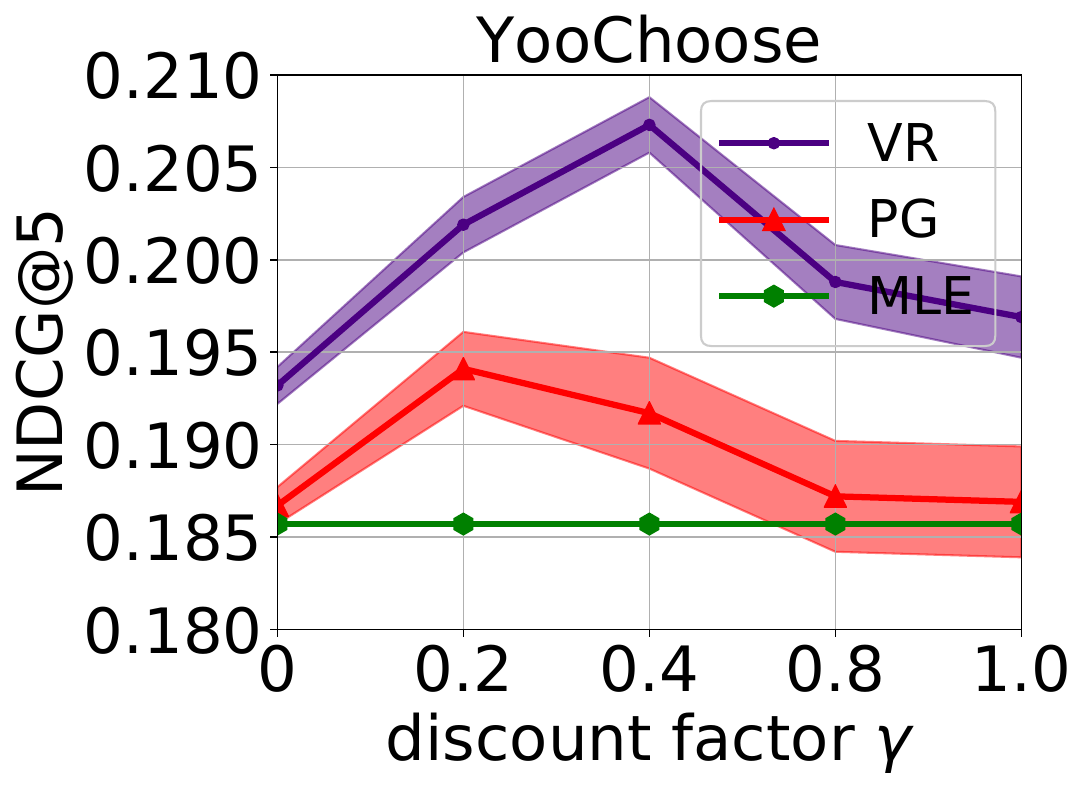}}
  \subfigure{
    \includegraphics[width=0.22\textwidth]{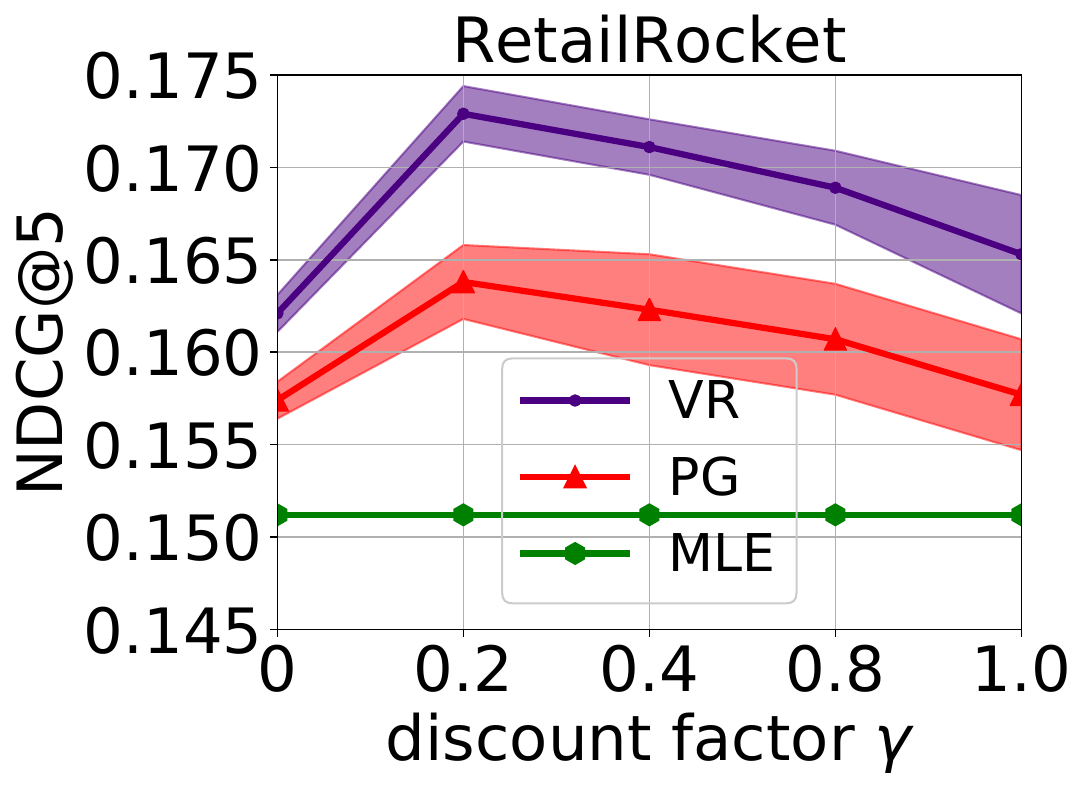}}
    \vskip -1.5em
    \caption{Performance with various discount factor $\gamma$.}\label{fig:discount}
\vskip -1.5em
\end{figure}

\begin{figure}[t]
\vskip -1em
\centering
  \subfigure{
    \includegraphics[width=0.22\textwidth]{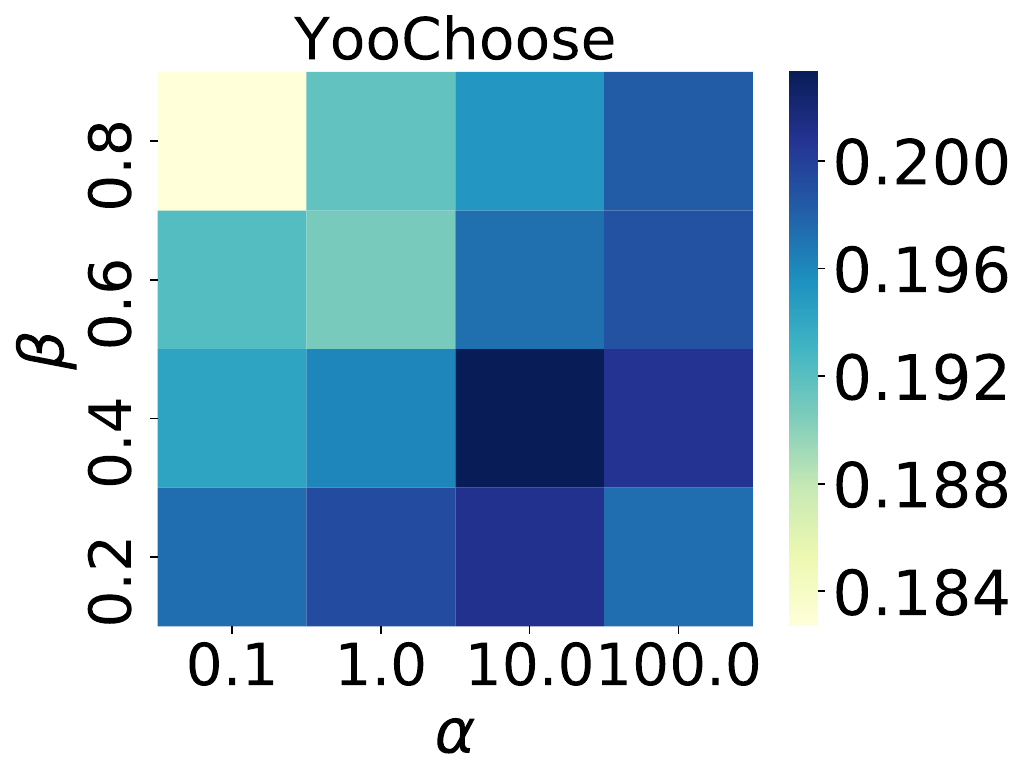}}
      \subfigure{
    \includegraphics[width=0.22\textwidth]{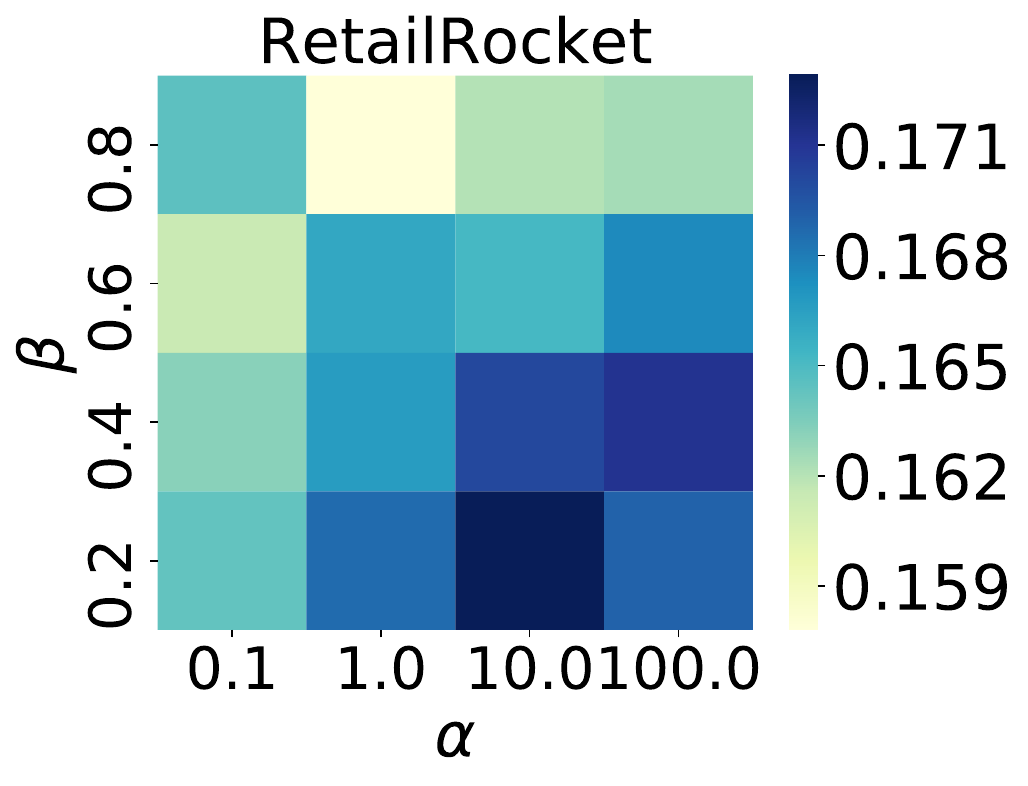}}
    \vskip -1.5em
    \caption{Performance (NDCG@5) with various $\alpha$ and $\beta$.}\label{fig:knowledge}
\vskip -1em
\end{figure}
\subsection{Further  Analysis}
 We take an  examination on VR to understand  how each component affects its performance. We follow the same setting in RQ1 and use SASRec as the backbone. \\
 \textbf{Future rewards}. We  investigate whether maximizing  future rewards  improves the ranking metric. In VR, the discount factor $\gamma$ controls the contribution of future rewards. For example, if  $\gamma=0$, VR is the bandit setting proposed in~\ref{Sec:One} and does not consider future rewards. Fig.~\ref{fig:discount} shows the performance  varying  $\gamma$. We can find that (1) appropriate values of $\gamma$ (maximizing future rewards)  can boost the performance but large
values  hurt the performance, and (2)  VR consistently performs better than baselines with every different $\gamma$.\\
\textbf{Knowledge transfer}. One of the most important properties of VR is transferring the knowledge between teacher and student. This transferring knowledge is controlled  by two key  parameters $\alpha$ and $\beta$. To understand their effects, we train VR  with various $\alpha$ and $\beta$. From~Fig.~\ref{fig:knowledge}, we can find smaller $\beta$ generally  leads better performance. This is because neglecting the ranking regularization term will result in the overestimation issue and lack of ranking signal. This is also consistent with our motivation. The performance  can be boosted by choosing appropriate values for $\alpha$ and $\beta$. \\
\textbf{Overestimation bias and variance}. As shown in theoretical analysis in \S~\ref{subsec:EV}, the estimation variance of the weighted TD loss in Eq.~(\ref{Eq:Q}) via  the importance sampling (IS) can be  reduced since our M-step Eq.~(\ref{Eq:VMM}) regularizes RL policy to the logged data via $\beta$. Table~\ref{Tbl:variance} shows the results of VR trained via TD loss and non-weighted TD loss. As shown in Table~\ref{Tbl:variance}, although IS introduces some variance,  VR can achieve a better bias-variance trade-off via adjusting  $\beta$. As shown in \S~\ref{sec:RLTR}, our VR can effectively reduce the overestimation bias. Fig.~\ref{fig:Overestimation} shows the overestimation bias on YooChoose and RetailRocket.  From Figs.~\ref{fig:pre} and ~\ref{fig:Overestimation}, we can find our VR achieves the best performance and can significantly reduce the overestimation bias compared with Q-learning, which empirically verifies our theoretic  analysis in \S~\ref{subsec:OB}.
\begin{table}[]
\center
\caption{Online evaluation results of CTR and coverage@3. Note that coverage  just indicates higher diversity and we list behavior policy just for reference, not for comparison.}
\vskip -1em
\resizebox{0.45\textwidth}{!}{
\begin{tabular}{@{}c ccc cc cccc c  @{}}
\toprule[1.0pt]
  & \multicolumn{2}{c}{Random}  & & \multicolumn{2}{c}{Maximum}   \\

  \cline{2-3}    \cline{5-6} 
     &  CTR  & Coverage$@$3   & &  CTR  & Coverage$@$3  &  \\
 
       \toprule[1.0pt]
     Pop     & 39.5 & 30.0  & &  41.8  & 30.0  &  \\
   MLE  &  63.5 $\pm$ 1.7  & 75.0 $\pm$ 1.8   & &  72.1 $\pm$  2.3  & 74.8 $\pm$ 2.1  &  \\
   DQN  &  50.7 $\pm$ 4.0  & 66.2 $\pm$ 2.3  & &  54.3 $\pm$ 2.1 & 68.4 $\pm$ 3.7  &  \\
    PG  &  72.5 $\pm$ 4.6 & 76.0 $\pm$ 2.9  & &  75.2 $\pm$ 3.8 & 76.5 $\pm$ 3.6  &  \\
      \toprule[0.7pt]
       	VR-NW  &  80.2 $\pm$ 1.2 & 78.5 $\pm$ 1.9   & & 78.9 $\pm$ 1.7 & 79.7 $\pm$ 0.9 &  \\
         VR  &  \textbf{84.3 $\pm$ 2.3} & \textbf{82.2 $\pm$ 2.7}   & &  \textbf{84.7 $\pm$ 3.3} & \textbf{85.6 $\pm$ 2.1} &  \\
                    \hline
        Behavior  &  63.1& 97.0  & & 75.2   & 74.5 &  \\ 
           
    \bottomrule[1.0pt]
\end{tabular}}
\label{Table:Online}
\vskip -1.5em
\end{table}

\subsection{Online Performance Comparison}
In this section, we conduct the online evaluation. We adopt RecSim~\cite{ie2019recsim} which is a simulation environment for testing RL for RS. We consider the default settings of RecSim with the number of categories as 20. We also consider logged data obtained from two behavioral policies: random (exploration)
policy and maximum reward policy which recommends the
top k items with the highest ground-truth reward.
From Table~\ref{Table:Online}, we can find: (1) DQN can not beat the behavior policy even the supervised MLE methods, which is consistent with our observations in the offline test.  This is because that DQN suffers from overestimation bias. (2) More importantly, our proposed VR significantly improves upon the behavior policy in CTR, which confirms that our methods can also perform well on online test in which there is an explicit distribution shift. (3) Our VR  significantly outperforms VR-NW, which shows the effectiveness of the importance sampling in the online testing.

%% file: Conclusion.tex
\section{Conclusion}
In this paper, we investigated offline learning of ranking policies for the sequential recommendation. We have presented a general EM framework, which can effectively learn from both rewards and ranking signals. To maximize long-term rewards, we extended the EM framework into sequential settings and proposed an off-policy VR algorithm. We provide theoretical guarantees of reducing overestimation bias and estimated variance. We show empirically that for the task of sequential recommendation, backbones learned by our VR outperform the same models that are optimized by other criteria such as MLE, off-policy PG, and DQN. 